\documentclass[letterpaper, 10 pt, conference]{ieeeconf}  

\IEEEoverridecommandlockouts                              

\usepackage{graphics} 
\usepackage{float}
\usepackage{epsfig} 
\usepackage{mathptmx} 
\usepackage{times} 
\usepackage{amsmath} 
\usepackage{amssymb}  
\usepackage{booktabs} 
\usepackage{multirow}
\usepackage{xcolor}
\usepackage{float} 
\usepackage{csquotes}
\usepackage[colorlinks=true, linkcolor=black, citecolor=black, urlcolor=blue]{hyperref}
\newcommand{\methodname}{SurgAM}

\newif\ifanonymous
\anonymoustrue 
\anonymousfalse

\title{\LARGE \bf
SurgAM: Surgical Affordance Map Prediction with\\Multimodal Feature Fusion for Robot Autonomy}
\ifanonymous
    \author{Anonymous Author}
\else
    \author{Lei Song, Yonghao Long, Mengya Xu, Jiayi Geng, Xiuyuan Chen$^{\dag}$, Qi Dou$^{\dag}$
    \thanks{L. Song, Y. Long, M. Xu, and Q. Dou are with the Department of Computer Science and Engineering, The Chinese University of Hong Kong.
    J. Geng and X. Chen are with the Department of Thoracic Surgery, Peking University People’s Hospital, Beijing, China.
    J. Geng and X. Chen are with the Thoracic Oncology Institute, Peking University People's Hospital, Beijing, China.
    J. Geng and X. Chen are with the Research Unit of Intelligence Diagnosis and Treatment in Early Non-small Cell Lung Cancer, Chinese Academy of Medical Sciences, 2021RU002, Peking University People’s Hospital, Beijing, China.
    J. Geng and X. Chen are with the Institute of Advanced Clinical Medicine, Peking University, Beijing, China.
    J. Geng and X. Chen are with Beijing Key Laboratory of Innovative Application of Big Data in Lung Cancer, Peking University People’s Hospital, Beijing, China.
    Corresponding authors: Qi Dou (qidou@cuhk.edu.hk), Xiuyuan Chen (dr\_chenxy@pku.edu.cn).
    }
    }
\fi

\begin{document}

\maketitle
\thispagestyle{empty}
\pagestyle{empty}

\begin{abstract}
Surgical automation is being increasingly studied, yet bridging visual scene understanding with autonomous action planning remains a fundamental challenge. While much research effort has been made on scene perception (e.g., tool recognition and scene segmentation), understanding and predicting actionable possibilities for surgical automation is still underexplored.
In this paper, we introduce surgical affordance prediction, which identifies actionable regions for fundamental surgical actions from visual data. Specifically, a novel adaptive feature fusion framework is proposed that leverages the complementary strengths of a self-supervised vision transformer encoder for its superior semantic understanding and a large-scale generative model encoder for its spatially-aware capability. Furthermore, we introduce a hierarchical prompt learning mechanism to adapt to varying procedural contexts. Finally, a scene-guided attention decoder is proposed to focus on critical surgical areas while suppressing background distractions. To validate the effectiveness, we established a new dataset, derived from publicly available surgical datasets with affordance annotations for three basic surgical actions: aspiration, clipping, and retraction.
Extensive experiments demonstrate that our approach achieves state-of-the-art performance.
Moreover, we validate our framework's applicability for downstream automation on a realistic lung and prostate phantom, and results show that the predicted affordance maps successfully enable autonomous surgical actions. 
\end{abstract}

\section{INTRODUCTION}
Robotic surgery has been increasingly adopted in modern healthcare, demonstrating promising clinical benefits.
Building on this success, there is a growing trend toward surgical automation~\cite{schmidgall2025will}, which promises to further reduce surgeon workload while improving procedural efficiency toward higher-level autonomy and human-robot collaboration~\cite{fiorini2022concepts}. 
To this end, autonomous robotic surgical systems must not only recognize what is in the surgical scene (e.g., surgical instruments or anatomical structure), but also interpret the scene in a way that directly supports manipulation, understanding what can be done where, and how to do it safely and effectively~\cite{attanasio2021autonomy}. Typical perception tasks, such as key point extraction, instrument detection, and scene segmentation~\cite{twinanda2017endonet}, have been widely studied, which provided valuable scene information. However, they just describe what is in the field of view, not where the robot should act~\cite{ayobi2025pixel}. 
The bridge between visual scene understanding and autonomous action planning remains a challenge for surgical automation~\cite{yang2017medical}. 

To address this challenge, traditional methods often rely on simplistic visual cues or static pre-operative models, which are insufficient for the highly dynamic scenes encountered in surgery. Recent attempts have explored different learning-based paradigms for surgical automation. For example, end-to-end approaches directly map visual inputs to robotic actions through imitation learning~\cite{kim2024surgical,moghani2025sufia}, demonstrating promise in controlled scenarios but suffering from overfitting to specific training demonstrations with poor generalization. Multistage learning methods decompose the problem into sequential modules, such as perception followed by planning and control~\cite{knudsen2024clinical}, where perception is often used for providing semantic mask and depth information, and control to plan the action. Although promising, these methods do not provide direct, interpretable guidance linking visual understanding to safe and effective actions across variable surgical contexts.


\begin{figure}[tp]
  \centering
  \includegraphics[scale=0.4]{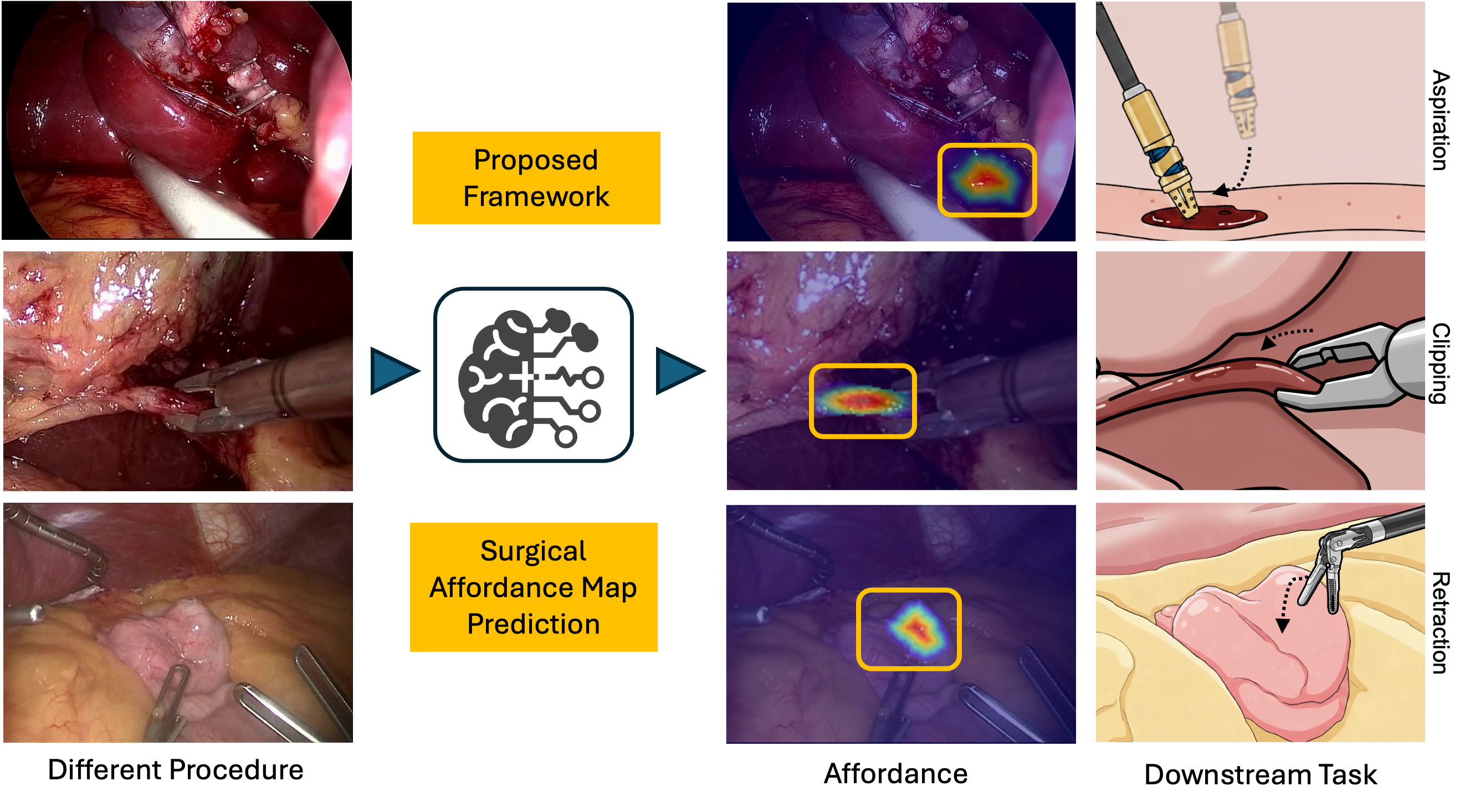}
  \vspace{-0.5em}
  \caption{Overall concept for surgical scene affordance map prediction: Model generates affordance map to identify optimal manipulation regions for specific surgical tasks, enabling downstream applications in robotic surgery and surgical planning.}
  \label{fig:teaser}
  \vspace{-0.6cm} 
\end{figure}

Affordance prediction in industrial robotics directly addresses such a challenge by predicting actionable possibilities of manipulating objects or environments. Specifically, affordance maps identify regions that support specific interactions, forming an interpretable bridge from perception to action~\cite{gibson2014ecological,ardon2020affordances}. For instance, grasping affordance maps highlight optimal contact points and orientations for a gripper~\cite{chen2024robotic}, while pushing affordance maps indicate where applied forces will yield desired object motions~\cite{zhao2024position}. These approaches are effective because they encode both geometric constraints and functional relationships between objects and actions, enabling generalization across objects and environments for robust robot manipulation.

Despite the success of affordance prediction in industrial robotics, its application for surgical robotics still remains unexplored. In this paper, we propose to predict the surgical affordance map: estimating spatial target regions suitable for fundamental surgical actions to provide direct, actionable guidance for surgical robotics. To the best of our knowledge, this is the first comprehensive study of surgical affordance map prediction. An illustration in Fig.~\ref{fig:teaser} shows affordance map prediction for fundamental surgical actions including aspiration, clipping, and retraction, where each map identifies optimal manipulation regions to guide robotic surgical interventions. Specifically, we propose a novel feature fusion framework that generates powerful visual representation by fusing dense semantic features with spatial features. Our key insight is that leveraging pre-trained vision models can provide complementary strengths for surgical affordance prediction. Specifically, transformers trained with self-supervised objectives demonstrate strong feature representation capabilities for fine-grained visual understanding, while generative models trained on diverse visual data excel at capturing spatial layout and providing smooth, coherent representations of anatomical structures. By combining these with an adaptive fusion strategy, hierarchical prompt learning, and a cross-modal alignment decoder, our framework generates robust and accurate surgical affordance maps.

Our main contributions are summarized as follows:
\begin{itemize}
\item We present the first study on affordance map prediction for surgical applications, bridging visual scene understanding and actionable robotic guidance.

\item We propose a new feature fusion framework leveraging semantic and spatial features with adaptive fusion, hierarchical prompt learning, and cross-modal alignment.

\item We establish a surgical affordance dataset with three fundamental surgical actions (aspiration, clipping, retraction) and demonstrate significant improvements over state-of-the-art baselines.

\item We validate our method's practical applicability through integration with an existing surgical automation framework and realistic phantom experiments on da Vinci Research Kit (dVRK).
\end{itemize}

\section{RELATED WORK} 

\subsection{Surgical Scene Perception}
Traditional surgical scene perception has focused on component-level understanding through specific subtasks such as instrument segmentation~\cite{shvets2018automatic}, anatomical structure detection~\cite{kitaguchi2020automated}, and surgical workflow recognition~\cite{czempiel2020tecno}. While these methods have achieved notable success in controlled scenarios, they provide isolated component identification that lacks holistic spatial-semantic reasoning necessary for autonomous surgical manipulation~\cite{attanasio2021autonomy}. Segmentation-based approaches can identify anatomical structures but fail to indicate where and how a robot should interact with these structures. Similarly, detection methods locate surgical instruments but provide limited guidance on feasible manipulation spaces within dynamic surgical environments~\cite{sen2016automating}. This limitation creates a critical gap between scene understanding and actionable intelligence, motivating affordance-based approaches for surgical manipulation.

\subsection{Autonomy in Robotic Surgery}
The pursuit of autonomy in robotic surgery aims to enhance precision, reduce surgeon fatigue, and standardize procedural outcomes~\cite{haidegger2019autonomy}. Early approaches often relied on extensive, task-specific programming for subtasks like suturing, which limited their adaptability~\cite{shademan2016supervised}. More recently, learning-based methods have demonstrated impressive capabilities. A landmark achievement by Kim et al. showed an autonomous robotic system successfully performing laparoscopic small bowel anastomosis in porcine models, outperforming human surgeons in consistency and accuracy~\cite{kim2025srt}. However, even in state-of-the-art systems, the perception-to-action pipeline often relies on simplified representations such as tracking fiducial markers or registering anatomy to a pre-operative plan~\cite{zhan2025tracking}. These methods typically lack a deep semantic understanding of the tissue itself, making them vulnerable to unexpected deformations or changes in the surgical scene. Explicitly identifying actionable regions as affordance maps remains largely unexplored, representing a critical missing link for robust surgical autonomy.

\subsection{Affordance Map Prediction}
Affordance prediction, rooted in ecological psychology~\cite{gibson2014ecological}, provides a natural bridge between perception and manipulation by identifying regions in an environment that support specific actions. In computer vision and robotics, this concept has evolved from early CNN-based methods~\cite{do2018affordancenet} to recent approaches that leverage large-scale foundation models. For instance, AffordanceLLM~\cite{qian2024affordancellm} utilizes the world knowledge of Vision Language Models for affordance grounding, and other works adapt models like CLIP for nuanced visual understanding~\cite{wang2023actionclip}. Concurrent research has also begun to develop foundation models specifically tailored for the surgical domain, demonstrating the growing interest in affordance prediction for robotic surgery. However, translating these general-domain methods to surgery remains challenging. Surgical scenes are characterized by deformable anatomy, occlusions, and dynamic lighting, which demand feature representations with both high semantic precision and spatial coherence~\cite{bommasani2021opportunities}. Recent work~\cite{zhang2023tale} has highlighted the complementary nature of features from models like Stable Diffusion~\cite{rombach2022high}, which excel in spatial layout, and DINOv2~\cite{oquab2023dinov2}, which provides robust semantic understanding. This suggests that a strategic fusion of such features holds significant potential for tackling the unique challenges of surgical affordance prediction.

\section{METHODS}

\begin{figure*}[ht]
  \centering
  \includegraphics[scale=0.77]{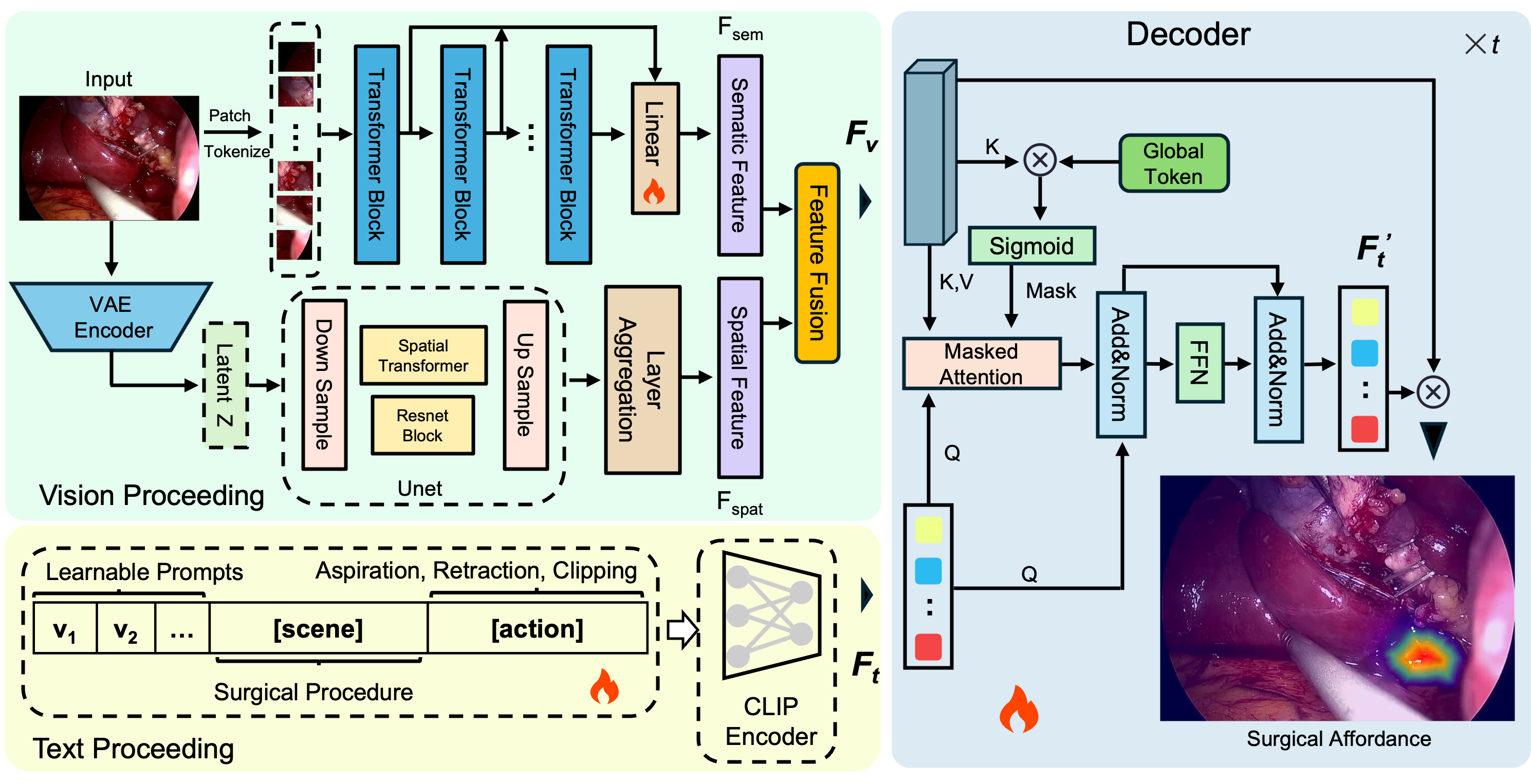}
  \vspace{-0.5em}
  \caption{Framework of surgical scene affordance map prediction. The architecture features two parallel streams on the left for multimodal feature extraction: a dual-vision encoder that fuses complementary features from a self-supervised vision transformer and a diffusion-based generative model, and a text encoder that leverages hierarchical prompt learning. On the right, a surgical scene guided cross-modal decoder integrates these features to generate the final affordance map.}
  \label{fig:framework}
  \vspace{-0.6cm} 
\end{figure*}

In this section, we present our novel framework for surgical scene affordance map prediction that leverages complementary foundation models and hierarchical prompt learning. Our approach addresses key challenges in surgical affordances understanding by combining the semantic precision of self-supervised vision transformers~\cite{oquab2023dinov2} with the spatial awareness of diffusion-based generative models~\cite{rombach2022high}, while incorporating surgical context through structured prompt learning to enhance domain-specific understanding.

\subsection{Overall Framework}
The prediction of surgical affordances in dynamic, deformable environments requires a visual representation that simultaneously captures two critical properties: fine-grained semantic precision and holistic spatial coherence. Semantic precision is vital for distinguishing between different anatomical structures, while spatial coherence is essential for defining smooth, physically actionable regions on tissue. The proposed learning framework, presented in Fig.~\ref{fig:framework}, is designed to meet this dual requirement through four key components: dual vision encoders, a complementary feature fusion module, a hierarchical text prompt encoder, and a surgical scene guided cross-modal decoder.

Central to our approach, two complementary vision encoders are used: one from a self-supervised vision transformer for dense semantic features, and the other from a diffusion-based generative model for spatially-aware representations. The fusion module then strategically combines these representations into a unified feature map. For text processing, we use a hierarchical prompt learning strategy to incorporate surgical context, enabling the model to distinguish between affordances in varying scenarios. Finally, a lightweight surgical scene guided cross-modal decoder generates affordance predictions by using a global context-guided attention mechanism to fuse the visual and text features, focusing on relevant surgical areas.

The following subsections detail each component's design and implementation, demonstrating how they collectively address the challenges of surgical affordance prediction.

\subsection{Complementary Visual Feature Extraction and Fusion}

A single type of pre-trained feature is often insufficient for the nuanced requirements of surgical affordance prediction~\cite{li2024review}. Recent studies have highlighted a fundamental trade-off in mainstream foundation models: features from generative models (diffusion models) provide spatially smooth but semantically imprecise representations, whereas features from self-supervised vision transformers are semantically accurate but often sparse and noisy~\cite{zhang2023tale}.

Therefore, our core technical contribution is a fusion strategy that synergizes these complementary feature types to form a more robust representation. We extract features celebrated for their semantic correspondence from a vision transformer~\cite{oquab2023dinov2} and features known for their spatial layout understanding from a diffusion-based generative model~\cite{rombach2022high}.

\noindent \textbf{Multi-Layer Semantic Feature Aggregation.}
To capture fine-grained semantic details at multiple granularity levels, we aggregate features from the last $j$ layers of the vision transformer:
\begin{equation}
\mathbf{F}_{sem} = \sum_{i=1}^{j} \alpha_i \mathbf{F}_i,
\end{equation}
where $\mathbf{F}_{sem}$ represents the semantic features, $\mathbf{F}_i$ denotes the $i$-th layer features, and the learnable weights with $\sum_i \alpha_i = 1$.

\noindent \textbf{Multi-Scale Spatial Feature Extraction.}
To capture the holistic spatial layout and context, we extract and concatenate features from multiple decoder layers of the generative model's U-Net:
\begin{equation}
\mathbf{F}_{spat} = \text{Concat}[\mathbf{F}_{spat}^{(2)}, \mathbf{F}_{spat}^{(5)}, \mathbf{F}_{spat}^{(8)}],
\end{equation}
where $\mathbf{F}_{spat}$ represents the spatial features from layers 2, 5, and 8, which are processed through appropriate dimensionality reduction.

\noindent \textbf{Adaptive Feature Fusion.}
To intelligently combine the strengths of both feature types, we implement an adaptive fusion strategy:
\begin{equation}
\mathbf{F}_v = \text{LayerNorm}(\beta \mathbf{F}_{sem} + (1-\beta) \mathbf{F}_{spat}),
\end{equation}
where $\beta$ is a learnable balancing parameter that is optimized during training to balance semantic precision and spatial coherence, and features are normalized before fusion to ensure training stability.

\subsection{Hierarchical Prompt Learning}

Manually designing prompts for surgical affordances presents significant challenges, as traditional templates~\cite{radford2021learning} like ``somewhere to [affordance]'' fail to capture the complex contextual information inherent in surgical procedures. For instance, ``clipping'' may refer to hemostatic control of active bleeding vessels, prophylactic sealing of at-risk vasculature, or tissue division for exposure—each requiring different spatial considerations and anatomical landmarks. Moreover, CLIP exhibits limited understanding of fine-grained affordances, particularly in the specialized domain of surgical robotics, where the same action may have different meanings across various procedural contexts.

To address these limitations, we extend the Context Optimization (CoOp)~\cite{zhou2022learning} method with a hierarchical prompt learning strategy:
\begin{equation}
\mathbf{P} = [\mathbf{v}_1, \ldots, \mathbf{v}_p, \mathbf{s}_{context}, \mathbf{s}_{action}],
\end{equation}
where $\mathbf{s}_{context}$ represents surgical context descriptors, $\{\mathbf{v}_i\}$ are learnable vectors, and $\mathbf{s}_{action}$ denotes the target affordance class.

The hierarchical prompts are processed through the CLIP text encoder to generate text embeddings:
\begin{equation}
\mathbf{F}_t = \text{CLIP}_{\text{text}}(\mathbf{P}),
\end{equation}
where $\mathbf{F}_t \in \mathbb{R}^{N \times D}$ represents the final text embeddings for $N$ affordance classes.
\subsection{Surgical Scene Guided Cross-Modal Attention}
The decoder takes as input the fused visual features $\mathbf{F}_v$ and text embeddings $\mathbf{F}_t$ to generate precise affordance predictions. Drawing inspiration from recent advances in vision-language alignment~\cite{li2024one}, we design a scene guided cross-attention mechanism that leverages global surgical context for precise affordance localization.

The attention mask is computed by projecting the global scene representation onto local visual features:
\begin{equation}
\mathbf{M} = \sigma\left(\frac{\mathbf{S}_{global} \mathbf{K}^T}{\sqrt{d}} \right),
\end{equation}
where $\mathbf{S}_{global}$ represents the global surgical scene token (CLS token) extracted from $\mathbf{F}_v$, and $\mathbf{K} = \text{Linear}(\mathbf{F}_v)$ denotes the key projections from visual features.

The masked cross-attention mechanism integrates text queries with visual representations:
\begin{equation}
\mathbf{F}_{out} = \text{softmax}\left(\frac{\mathbf{Q}\mathbf{K}^T}{\sqrt{d}} \odot \mathbf{M}\right) \mathbf{V} + \mathbf{F}_t,
\end{equation}
where $\mathbf{Q} = \text{Linear}(\mathbf{F}_t)$, $\mathbf{V} = \text{Linear}(\mathbf{F}_v)$, and the residual connection preserves the original text information.

The final surgical affordance prediction combines the refined embeddings:
\begin{equation}
\mathbf{P}_{aff} = \text{softmax}\left(\mathbf{F}_{out} \mathbf{F}_{v}^T\right),
\end{equation}
where $\mathbf{P}_{aff} \in \mathbb{R}^{N \times L}$ represents the spatial probability distribution over $N$ surgical actions across $L$ image locations.

\section{EXPERIMENTS}
In this section, we conduct a series of experiments to systematically evaluate our proposed framework. We aim to answer three key questions: (1) How does our full model perform against state-of-the-art methods on our challenging surgical affordance dataset? (2) What is the individual contribution of each core component of our design, particularly the dual-feature fusion strategy? (3) Can the predicted affordance maps be effectively translated into actionable guidance for a physical robot in surgical manipulation tasks?

\subsection{Dataset}\label{subsec:dataset}

To address the scarcity of specialized surgical data, we constructed a new affordance dataset from a collection of diverse, publicly available surgical video corpora. Our annotation methodology is uniquely designed to capture surgical \textit{intent}. For each interaction, we identify the ``pre-contact'' frame just before the tool touches the tissue and annotate a single point indicating the intended interaction location. This point then serves as a visual prompt for the Segment Anything Model (SAM)~\cite{kirillov2023segment} to generate a high-fidelity segmentation mask of the actionable and safe tissue region, which is subsequently converted into a heatmap to serve as the ground truth.

Our final dataset consists of annotations from 1,915 surgical video sequences from 7 public datasets: AutoLaparo~\cite{wang2022autolaparo}, CholecT50~\cite{nwoye2022rendezvous}, 
HeiChole~\cite{maier2021heidelberg}, MultiBypass140~\cite{lavanchy2024challenges}, SurgicalActions160~\cite{schoeffmann2018video}, Endovis18~\cite{allan20202018}, and MESAD-Real~\cite{bawa2021saras}, categorized into three fundamental affordance types: \textbf{Retraction}, \textbf{Clipping}, and \textbf{Aspiration}. The distribution of these clips is as follows: 904 for Retraction, 231 for Clipping, and 780 for Aspiration. This imbalanced distribution is representative of the natural frequency of these actions during actual surgical procedures, where tissue retraction and aspiration are more common maneuvers than clipping. This dataset provides a robust foundation for training and evaluating affordance prediction models in a variety of surgical contexts.

\begin{figure*}[ht]
  \centering
  \includegraphics[scale=0.75]{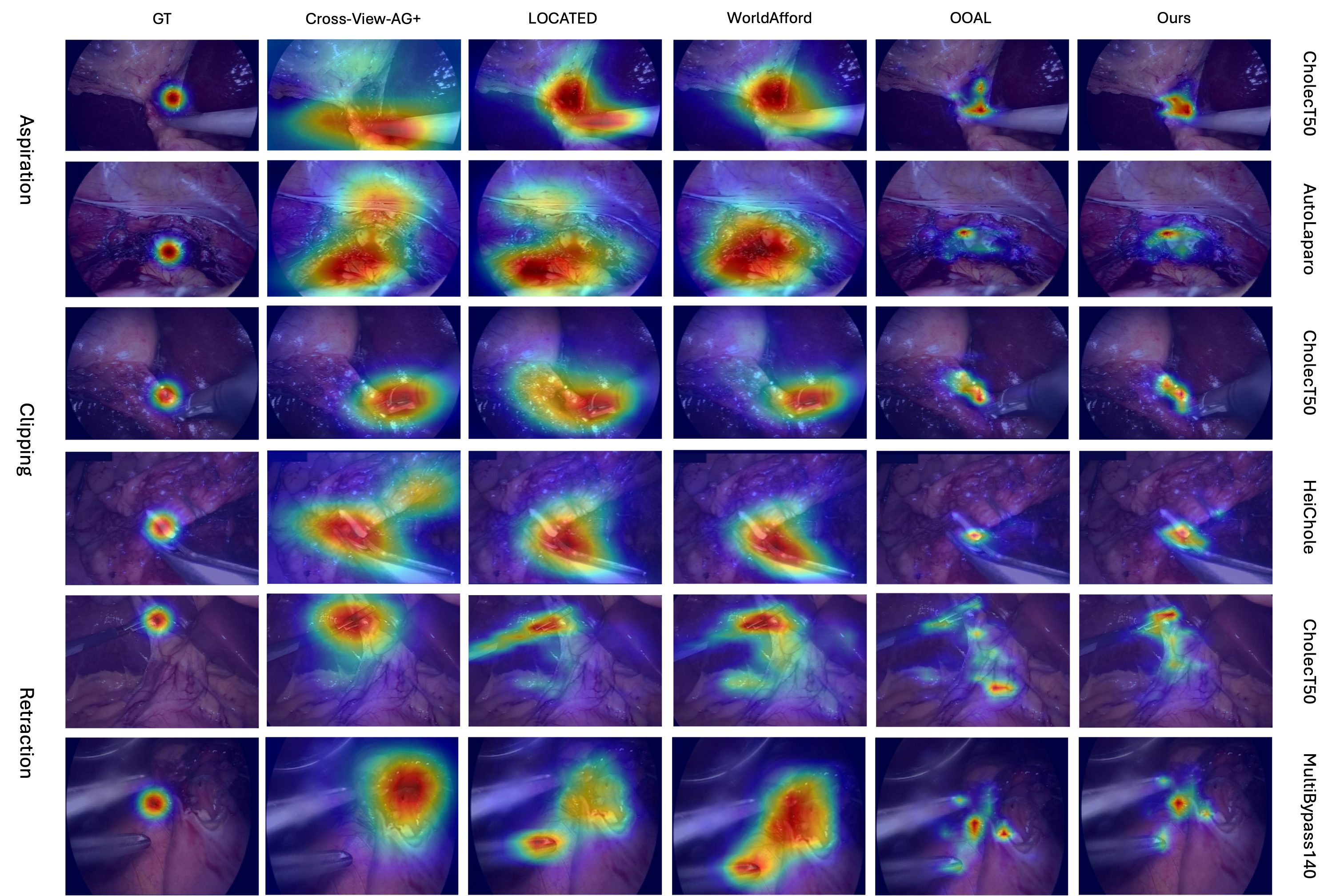}
  \vspace{-0.5em}
  \caption{Qualitative comparison of affordance prediction methods across different surgical tasks (aspiration, clipping, retraction) and datasets showing baseline methods vs. our approach}
  \label{fig:results}
  \vspace{-0.6cm} 
\end{figure*}

\subsection{Experimental Setup}
\noindent \textbf{Metrics.}
We evaluate our model on our affordance dataset (details in Sec.~\ref{subsec:dataset}) using a comprehensive set of established affordance prediction metrics~\cite{li2023locate, qian2023understanding}. To assess the similarity between the predicted and ground-truth probability distributions, we use the Kullback-Leibler Divergence (KLD), where a lower value indicates less divergence and a better match. To measure spatial overlap, we use the Similarity (SIM) metric, for which higher is better. We also employ Normalized Scanpath Saliency (NSS), which evaluates the model's ability to assign high values to the ground truth affordance regions; a higher NSS score signifies a more accurate prediction. Furthermore, we propose Centroid Localization Accuracy (CLA) to quantify spatial precision in a normalized manner. Unlike raw Average Centroid Distance (ACD), which depends on image resolution, CLA normalizes by the image diagonal D: $CLA = 1 - {ACD}/{D}$. This produces an interpretable score between 0 and 1, where higher values indicate better spatial accuracy.

\noindent \textbf{Implementation Details.}
We implement our framework using DINOv2~\cite{oquab2023dinov2} as the self-supervised vision transformer and Stable Diffusion~\cite{rombach2022high} as the diffusion-based generative model. Features are extracted using pre-trained checkpoints without fine-tuning backbone networks. For DINOv2, we aggregate features from the last 4 layers (j=4) with learnable weights. For Stable Diffusion, we extract features from U-Net decoder layers 2, 5, and 8, applying linear projections to align dimensions before fusion. The adaptive fusion parameter $\beta$ is initialized to 0.5 and learned during training.

\begin{table}[h]
   \centering
   \caption{\textbf{Quantitative results on the whole surgical dataset.}}
   \label{tab:main_easy}
   \vspace{-0.6em}
   \begin{tabular*}{0.95\columnwidth}{@{\extracolsep{\fill}}llcccc@{}}
      \toprule
      Methods & KLD $\downarrow$ & SIM $\uparrow$ & NSS $\uparrow$ & CLA $\uparrow$\\
      \midrule
      Cross-View-AG~\cite{luo2022learning}       & 1.795 & 0.260 & 1.227 & 0.765  \\
      Cross-View-AG+~\cite{luo2024grounded}      & 1.793 & 0.253 & 1.265 & 0.782  \\
      LOCATE~\cite{li2023locate}                 & 1.696 & 0.295 & 1.276 & 0.881  \\
      WorldAfford~\cite{chen2024worldafford} & 1.559 & 0.307 & 1.446 & 0.883  \\
      OOAL~\cite{li2024one}                      & 1.665 & 0.274 & 1.425 & 0.884  \\
      \textbf{\methodname\ (Ours)} & \textbf{1.362} & \textbf{0.367} & \textbf{1.642} & \textbf{0.895}  \\
      \bottomrule
   \end{tabular*}
   \vspace{-0.6em}
\end{table}

\subsection{Qualitative results and Analysis}
Fig.~\ref{fig:results} shows qualitative results on our test datasets. We compare our approach \methodname, with several state-of-the-art methods that learn affordances from 2D images. All baseline models are trained and tested on the same datasets to ensure a fair comparison.

On the surgical test datasets, we observe distinct failure patterns across different baseline methods, highlighting the architectural limitations in surgical affordance prediction. \textbf{Cross-View-AG}~\cite{luo2022learning} and \textbf{Cross-View-AG+}~\cite{luo2024grounded} produce overly 
expansive affordance maps due to their ACP strategy, despite 
achieving relatively high CLA scores (0.765-0.782). \textbf{LOCATE}~\cite{li2023locate} shows improved localization through its PartSelect mechanism, yet its k-means clustering approach creates instability in complex scenes, leading to extensive anatomical coverage as a conservative recall strategy (CLA: 0.881). \textbf{WorldAfford}~\cite{chen2024worldafford} leverages LLM reasoning through ARCoT but suffers from attention diffusion in its WCB module, producing over-activation patterns despite strong metrics (KLD: 1.559, CLA: 0.883). \textbf{OOAL}~\cite{li2024one} presents distinctly different behavior with more concentrated predictions due to its DINOv2-based architecture and few-shot learning strategy, avoiding overfitting but struggling with complex multi-object scenarios due to single-modality limitations.

Our approach \textbf{\methodname} consistently produces superior affordance maps that achieve an optimal balance between spatial precision and coverage completeness. Unlike baseline methods that achieve high confidence through over-activation strategies, our framework demonstrates genuine spatial accuracy with concentrated predictions that correspond to actual affordance regions. The synergistic fusion of Stable Diffusion's spatial understanding with DINOv2's semantic precision, enhanced by our surgical-aware hierarchical prompting strategy, enables accurate localization while maintaining spatial coherence. The substantial improvements across all evaluation metrics (KLD: 1.362, SIM: 0.367, NSS: 1.642, CLA: 0.895) validate the effectiveness of our dual-foundation model fusion approach. Notably, our method achieves the best KLD and NSS scores while maintaining competitive CLA performance, demonstrating that superior affordance prediction stems from architectural design rather than conservative prediction strategies employed by competing methods.

\begin{table}[H]
   \centering
   \caption{\textbf{Ablation results of different visual foundation models.}}
   \label{tab:abla1}
   \vspace{-0.6em}
   \scalebox{0.95}{
   \begin{tabular*}{0.95\columnwidth}{@{\extracolsep{\fill}}lcccc@{}}
      \toprule
      Model & KLD $\downarrow$ & SIM $\uparrow$ & NSS $\uparrow$ & CLA $\uparrow$\\
      \midrule
      CLIP & 2.660 & 0.156 & 0.276 & 0.819 \\
      DeiT III & 2.231 & 0.165 & 0.292 & 0.531 \\
      SD & 2.242 & 0.187 & 0.542 & 0.834 \\
      DINOv2 & 1.808 & 0.237 & 1.313 & 0.837 \\
      \bottomrule
   \end{tabular*}
   } %
   \vspace{-0.6em}
\end{table}

\begin{table}[H] 
    \centering
    \caption{\textbf{Ablation results of the proposed modules.} HPL: Hierarchical Prompt Learning. SD-DINO: Stable Diffusion and DINOv2 feature fusion. TD: Transformer Decoder. SGM: Surgical Scene Guided Mask}
    \vspace{-0.6em}
    \label{tab:abla2}
    \scalebox{0.95}{
    \begin{tabular*}{\columnwidth}{@{\extracolsep{\fill}}cccc cccc@{}}
        \toprule
        \multicolumn{4}{c}{\textbf{Components}} & \multicolumn{4}{c}{\textbf{Metrics}} \\
        \cmidrule(r){1-4} \cmidrule(l){5-8} 
        HPL & SD-DINO & TD & SGM & KLD $\downarrow$ & SIM $\uparrow$ & NSS $\uparrow$ & CLA $\uparrow$\\
        \midrule
         & & & & 1.808 & 0.237 & 1.313 & 0.837 \\
        \checkmark & & & & 1.742 & 0.268 & 1.240 & 0.847 \\
        \checkmark & \checkmark & & & 1.473 & 0.334 & 1.528 & 0.885 \\
        \checkmark & \checkmark & \checkmark & & 1.432 & 0.348 & 1.552 & 0.892 \\
        \checkmark & \checkmark & \checkmark & \checkmark & 1.362 & 0.367 & 1.642 & 0.895  \\
        \bottomrule
    \end{tabular*}
    } 
    \vspace{-0.6em}
\end{table}

\subsection{Ablation study}
\label{sec:ablation}
We conduct comprehensive ablation studies to validate the effectiveness of each component in our framework. The experiments are performed on our whole dataset, which encompasses diverse surgical scenarios and complex anatomical structures.
\noindent\textbf{Visual Foundation Model Comparison.}
Table~\ref{tab:abla1} presents a systematic comparison of different visual foundation models for surgical affordance prediction. CLIP and DeiT III demonstrate limited effectiveness in surgical contexts, with poor KLD (2.660, 2.231) and NSS (0.276, 0.292) scores, indicating substantial distribution mismatch and weak affordance localization capabilities. Stable Diffusion features exhibit superior spatial understanding, achieving better NSS (0.542) and CLA (0.834) scores, validating that SD's spatial layout training provides valuable coherence for surgical scenes. However, SD underperforms in semantic precision (KLD: 2.242). DINOv2 demonstrates clear superiority across all metrics (KLD: 1.808, NSS: 1.313), with nearly 5× better NSS than CLIP. This superior performance stems from self-supervised training that learns part-aware representations, making it well-suited for fine-grained spatial understanding required in surgical affordance prediction.

\noindent\textbf{Component-wise Analysis.}
Table~\ref{tab:abla2} demonstrates the progressive improvement achieved by incorporating each proposed component. Starting from the DINOv2 baseline (KLD: 1.808), each module contributes meaningful performance gains, culminating in substantial improvements in the full model (KLD: 1.362, NSS: 1.642).

The most significant breakthrough comes from the SD-DINO fusion module, which represents the core innovation of our approach. This fusion strategy addresses the fundamental limitations observed in single-modality approaches by combining SD's spatial coherence with DINOv2's semantic precision. The substantial performance improvement validates our central hypothesis that surgical affordance prediction requires both spatial layout understanding and semantic correspondence capabilities, which qualities neither foundation model possesses independently.

The Hierarchical Prompt Learning (HPL) component provides essential domain adaptation, enabling the model to distinguish between affordances across different surgical contexts. The Transformer Decoder (TD) enhances contextual modeling by enabling sophisticated feature interactions, while the Surgical Scene Guided Mask (SGM) provides crucial attention focusing by leveraging global surgical context to suppress background distractions.

The final integrated system achieves optimal performance across all metrics, with particularly notable improvements in NSS (25.0\% gain) and substantial KLD reduction (24.7\% improvement), demonstrating that the synergistic combination of all components successfully addresses the multifaceted challenges of surgical affordance prediction.

\subsection{Deployment and Validation of Autonomy on Phantom}
The ablation studies confirmed the effectiveness of our design choices in generating accurate surgical affordance maps. To further validate the practical utility, we integrated our affordance prediction module with a visual servoing controller based on our VPPV framework~\cite{long2025surgical} and deployed the overall framework on a physical surgical robot platform  to perform surgical tasks autonomously. 

\begin{figure}[H]
\centering
\includegraphics[width=0.46\textwidth]{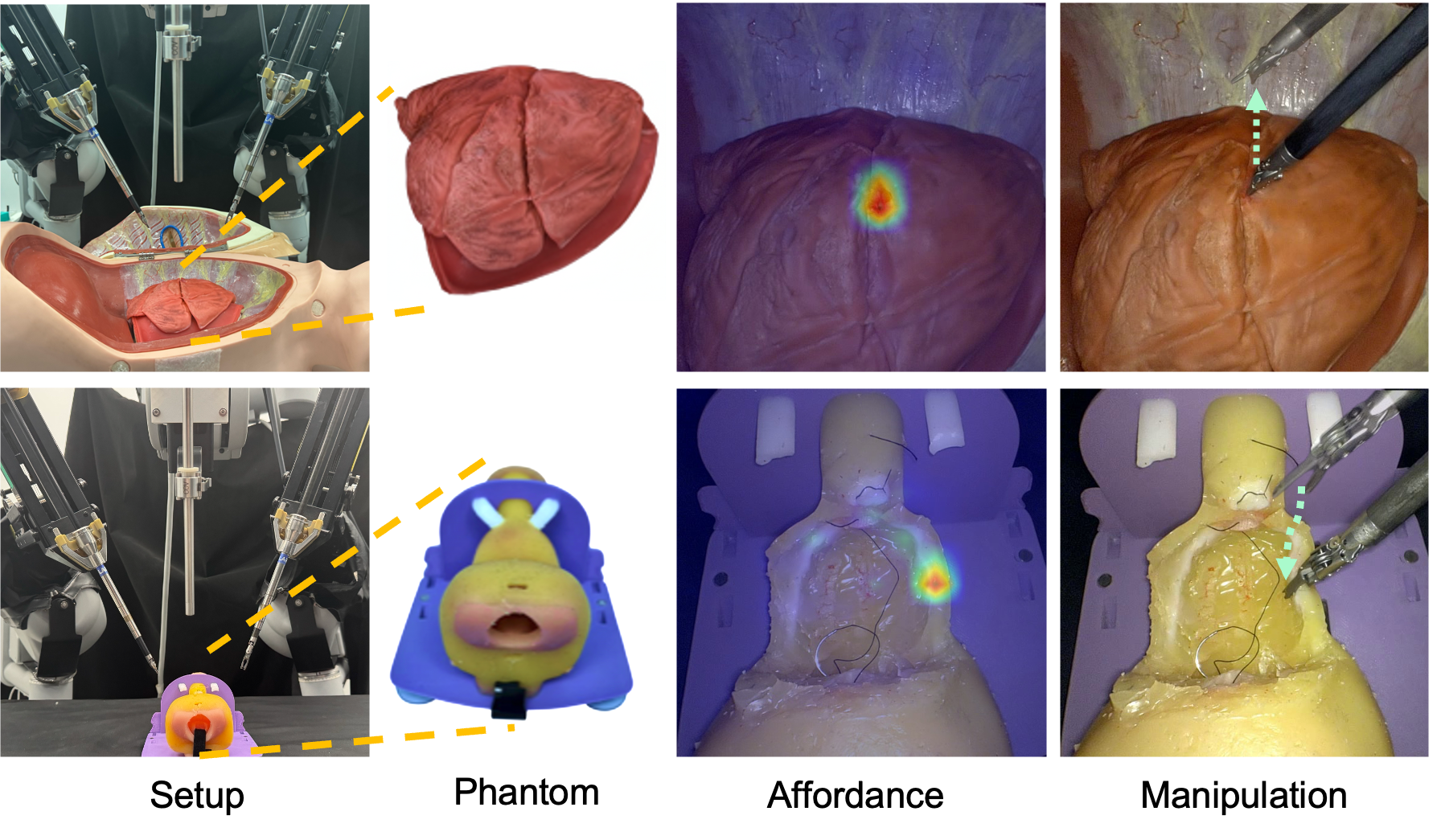}
\vspace{-0.6em}
\caption{Experimental setup for autonomous task validation using dVRK. Each row shows (from left to right): dVRK setup, phantom model, affordance prediction, and autonomous manipulation for lung retraction (top) and prostate clipping (bottom) tasks.}
\label{fig:phantom}
\vspace{-0.5em} 
\end{figure}

\noindent\textbf{Experimental Setup.}
Our physical setup, shown in Fig.~\ref{fig:phantom}, consists of a da Vinci Research Kit (dVRK) and two high-fidelity surgical phantoms: a torso phantom with a silicone lung model and a standalone prostate phantom with simulated vasculature. The dVRK is equipped with an Endoscope Camera Manipulator (ECM) for visual feedback and two Patient Side Manipulators (PSMs) to control surgical instruments, enabling us to replicate key visual and kinematic aspects of different surgical scenarios. For each phantom, we collected targeted datasets through teleoperated demonstrations and fine-tuned task-specific models while retaining generalized features. During autonomous execution, our framework processed real-time video to generate affordance maps that guided the visual servoing controller.

\noindent\textbf{Autonomous Task Execution and Results.}
During the autonomous execution, our framework processed real-time video feeds from the ECM to generate affordance maps, which guide a visual servoing controller that commanded the PSM to perform the designated manipulation. For autonomous lung retraction with simulated circulation dynamics, our system consistently identified optimal retraction regions (as shown in the first row of Fig.\ref{fig:phantom}) while navigating tissue movement and deformation, achieving $100\%$ success rate across 50 trials. For autonomous vessel clipping on the prostate phantom, we evaluated a clinically relevant workflow where expert surgeons first exposed target vessels through teleoperation, followed by autonomous clipping execution (prediction result shown in the bottom row of Fig.\ref{fig:phantom}), achieving $98\%$ success rate (49/50 trials). The single failure occurred when camera repositioning temporarily occluded the target vessel, but affordance prediction immediately recovered once visibility was restored, demonstrating robust recovery capabilities. These results across two distinct surgical scenarios demonstrate that our framework provides reliable perception-to-action guidance for robotic surgical manipulation. However, complete surgical autonomy requires integration with comprehensive control systems and validation beyond controlled phantom environments.

\section{CONCLUSION AND DISCUSSION}
This paper presents a novel framework for surgical scene affordance map prediction that leverages the complementary strengths of diffusion-based generative model and vision transformer encoder. Our adaptive fusion strategy combines SD's spatial coherence with DINOv2's semantic precision, enhanced by hierarchical prompt learning and scene-guided attention mechanisms to address the unique challenges of surgical environments. Extensive experiments demonstrate the effectiveness of our approach across diverse surgical scenarios, with successful phantom implementations validating the practical applicability for autonomous surgical manipulation tasks.
This work represents a significant step toward intelligent surgical robotic systems capable of adaptive decision-making in complex surgical environments. Future work will focus on incorporating temporal dynamics, expanding the annotated dataset scope, and conducting comprehensive validation studies on live surgical procedures.

\section*{ACKNOWLEDGMENT}
This work was supported in part by a grant from the Research 
Grants Council of the Hong Kong Special Administrative Region, 
China (Project No. 14208424), and in part by the InnoHK 
initiative of the Innovation and Technology Commission of the 
Hong Kong Special Administrative Region Government. 
\par
The work from J. Geng and X. Chen has been supported in part by the 
Peking University Medicine Plus X Pilot Program-Artificial 
Intelligence and Medical Development Initiative 
(BMU2025YXXLHAIYX002), Peking University People's Hospital 
Scientific Research Development Funds (RDX2024-07), and S\&T 
Program of Xiongan New Area (XA202501102003K). The authors 
would like to thank Yiru Ye, Oranuch Ekkowit, Phutanate Pisutsin and Wong Pak Hin for their valuable contributions to the dataset annotation.

\bibliographystyle{IEEEtran}
\bibliography{IEEEabrv,references}

@book{gibson2014ecological,
  title={The ecological approach to visual perception: classic edition},
  author={Gibson, James J},
  year={2014},
  publisher={Psychology press}
}

@article{haidegger2019autonomy,
  title={Autonomy for surgical robots: Concepts and paradigms},
  author={Haidegger, Tam{\'a}s},
  journal={IEEE Transactions on Medical Robotics and Bionics},
  volume={1},
  number={2},
  pages={65--76},
  year={2019},
  publisher={IEEE}
}

@inproceedings{li2023locate,
  title={Locate: Localize and transfer object parts for weakly supervised affordance grounding},
  author={Li, Gen and Jampani, Varun and Sun, Deqing and Sevilla-Lara, Laura},
  booktitle={Proceedings of the IEEE/CVF Conference on Computer Vision and Pattern Recognition},
  pages={10922--10931},
  year={2023}
}

@inproceedings{luo2022learning,
  title={Learning affordance grounding from exocentric images},
  author={Luo, Hongchen and Zhai, Wei and Zhang, Jing and Cao, Yang and Tao, Dacheng},
  booktitle={Proceedings of the IEEE/CVF conference on computer vision and pattern recognition},
  pages={2252--2261},
  year={2022}
}

@inproceedings{qian2023understanding,
  title={Understanding 3d object interaction from a single image},
  author={Qian, Shengyi and Fouhey, David F},
  booktitle={Proceedings of the IEEE/CVF International Conference on Computer Vision},
  pages={21753--21763},
  year={2023}
}

@article{luo2024grounded,
  title={Grounded affordance from exocentric view},
  author={Luo, Hongchen and Zhai, Wei and Zhang, Jing and Cao, Yang and Tao, Dacheng},
  journal={International Journal of Computer Vision},
  volume={132},
  number={6},
  pages={1945--1969},
  year={2024},
  publisher={Springer}
}

@inproceedings{li2024one,
  title={One-shot open affordance learning with foundation models},
  author={Li, Gen and Sun, Deqing and Sevilla-Lara, Laura and Jampani, Varun},
  booktitle={Proceedings of the IEEE/CVF Conference on Computer Vision and Pattern Recognition},
  pages={3086--3096},
  year={2024}
}

@inproceedings{rombach2022high,
  title={High-resolution image synthesis with latent diffusion models},
  author={Rombach, Robin and Blattmann, Andreas and Lorenz, Dominik and Esser, Patrick and Ommer, Bj{\"o}rn},
  booktitle={Proceedings of the IEEE/CVF conference on computer vision and pattern recognition},
  pages={10684--10695},
  year={2022}
}

@article{oquab2023dinov2,
  title={Dinov2: Learning robust visual features without supervision},
  author={Oquab, Maxime and Darcet, Timoth{\'e}e and Moutakanni, Th{\'e}o and Vo, Huy and Szafraniec, Marc and Khalidov, Vasil and Fernandez, Pierre and Haziza, Daniel and Massa, Francisco and El-Nouby, Alaaeldin and others},
  journal={arXiv preprint arXiv:2304.07193},
  year={2023}
}

@article{zhou2022learning,
  title={Learning to prompt for vision-language models},
  author={Zhou, Kaiyang and Yang, Jingkang and Loy, Chen Change and Liu, Ziwei},
  journal={International Journal of Computer Vision},
  volume={130},
  number={9},
  pages={2337--2348},
  year={2022},
  publisher={Springer}
}

@inproceedings{radford2021learning,
  title={Learning transferable visual models from natural language supervision},
  author={Radford, Alec and Kim, Jong Wook and Hallacy, Chris and Ramesh, Aditya and Goh, Gabriel and Agarwal, Sandhini and Sastry, Girish and Askell, Amanda and Mishkin, Pamela and Clark, Jack and others},
  booktitle={International conference on machine learning},
  pages={8748--8763},
  year={2021},
  organization={PmLR}
}

@article{zhang2023tale,
  title={A tale of two features: Stable diffusion complements dino for zero-shot semantic correspondence},
  author={Zhang, Junyi and Herrmann, Charles and Hur, Junhwa and Polania Cabrera, Luisa and Jampani, Varun and Sun, Deqing and Yang, Ming-Hsuan},
  journal={Advances in Neural Information Processing Systems},
  volume={36},
  pages={45533--45547},
  year={2023}
}

@inproceedings{qian2024affordancellm,
  title={Affordancellm: Grounding affordance from vision language models},
  author={Qian, Shengyi and Chen, Weifeng and Bai, Min and Zhou, Xiong and Tu, Zhuowen and Li, Li Erran},
  booktitle={Proceedings of the IEEE/CVF Conference on Computer Vision and Pattern Recognition},
  pages={7587--7597},
  year={2024}
}

@inproceedings{do2018affordancenet,
  title={Affordancenet: An end-to-end deep learning approach for object affordance detection},
  author={Do, Thanh-Toan and Nguyen, Anh and Reid, Ian},
  booktitle={2018 IEEE international conference on robotics and automation (ICRA)},
  pages={5882--5889},
  year={2018},
  organization={IEEE}
}

@article{ardon2020affordances,
  title={Affordances in robotic tasks--a survey},
  author={Ard{\'o}n, Paola and Pairet, {\`E}ric and Lohan, Katrin S and Ramamoorthy, Subramanian and Petrick, Ronald},
  journal={arXiv preprint arXiv:2004.07400},
  year={2020}
}

@inproceedings{kirillov2023segment,
  title={Segment anything},
  author={Kirillov, Alexander and Mintun, Eric and Ravi, Nikhila and Mao, Hanzi and Rolland, Chloe and Gustafson, Laura and Xiao, Tete and Whitehead, Spencer and Berg, Alexander C and Lo, Wan-Yen and others},
  booktitle={Proceedings of the IEEE/CVF international conference on computer vision},
  pages={4015--4026},
  year={2023}
}

@article{wang2023actionclip,
  title={Actionclip: Adapting language-image pretrained models for video action recognition},
  author={Wang, Mengmeng and Xing, Jiazheng and Mei, Jianbiao and Liu, Yong and Jiang, Yunliang},
  journal={IEEE Transactions on Neural Networks and Learning Systems},
  year={2023},
  publisher={IEEE}
}

@misc{yang2017medical,
  title={Medical robotics—Regulatory, ethical, and legal considerations for increasing levels of autonomy},
  author={Yang, Guang-Zhong and Cambias, James and Cleary, Kevin and Daimler, Eric and Drake, James and Dupont, Pierre E and Hata, Nobuhiko and Kazanzides, Peter and Martel, Sylvain and Patel, Rajni V and others},
  journal={Science robotics},
  volume={2},
  number={4},
  pages={eaam8638},
  year={2017},
  publisher={American Association for the Advancement of Science}
}

@inproceedings{sen2016automating,
  title={Automating multi-throw multilateral surgical suturing with a mechanical needle guide and sequential convex optimization},
  author={Sen, Siddarth and Garg, Animesh and Gealy, David V and McKinley, Stephen and Jen, Yiming and Goldberg, Ken},
  booktitle={2016 IEEE international conference on robotics and automation (ICRA)},
  pages={4178--4185},
  year={2016},
  organization={IEEE}
}

@article{shademan2016supervised,
  title={Supervised autonomous robotic soft tissue surgery},
  author={Shademan, Azad and Decker, Ryan S and Opfermann, Justin D and Leonard, Simon and Krieger, Axel and Kim, Peter CW},
  journal={Science translational medicine},
  volume={8},
  number={337},
  pages={337ra64--337ra64},
  year={2016},
  publisher={American Association for the Advancement of Science}
}

@article{attanasio2021autonomy,
  title={Autonomy in surgical robotics},
  author={Attanasio, Aleks and Scaglioni, Bruno and De Momi, Elena and Fiorini, Paolo and Valdastri, Pietro},
  journal={Annual Review of Control, Robotics, and Autonomous Systems},
  volume={4},
  number={1},
  pages={651--679},
  year={2021},
  publisher={Annual Reviews}
}

@article{bommasani2021opportunities,
  title={On the opportunities and risks of foundation models},
  author={Bommasani, Rishi and Hudson, Drew A and Adeli, Ehsan and Altman, Russ and Arora, Simran and von Arx, Sydney and Bernstein, Michael S and Bohg, Jeannette and Bosselut, Antoine and Brunskill, Emma and others},
  journal={arXiv preprint arXiv:2108.07258},
  year={2021}
}

@inproceedings{shvets2018automatic,
  title={Automatic instrument segmentation in robot-assisted surgery using deep learning},
  author={Shvets, Alexey A and Rakhlin, Alexander and Kalinin, Alexandr A and Iglovikov, Vladimir I},
  booktitle={2018 17th IEEE international conference on machine learning and applications (ICMLA)},
  pages={624--628},
  year={2018},
  organization={IEEE}
}

@article{kitaguchi2020automated,
  title={Automated laparoscopic colorectal surgery workflow recognition using artificial intelligence: experimental research},
  author={Kitaguchi, Daichi and Takeshita, Nobuyoshi and Matsuzaki, Hiroki and Oda, Tatsuya and Watanabe, Masahiko and Mori, Kensaku and Kobayashi, Etsuko and Ito, Masaaki},
  journal={International journal of surgery},
  volume={79},
  pages={88--94},
  year={2020},
  publisher={Elsevier}
}

@inproceedings{czempiel2020tecno,
  title={Tecno: Surgical phase recognition with multi-stage temporal convolutional networks},
  author={Czempiel, Tobias and Paschali, Magdalini and Keicher, Matthias and Simson, Walter and Feussner, Hubertus and Kim, Seong Tae and Navab, Nassir},
  booktitle={Medical Image Computing and Computer Assisted Intervention--MICCAI 2020: 23rd International Conference, Lima, Peru, October 4--8, 2020, Proceedings, Part III 23},
  pages={343--352},
  year={2020},
  organization={Springer}
}

@article{fiorini2022concepts,
  title={Concepts and trends in autonomy for robot-assisted surgery},
  author={Fiorini, Paolo and Goldberg, Ken Y and Liu, Yunhui and Taylor, Russell H},
  journal={Proceedings of the IEEE},
  volume={110},
  number={7},
  pages={993--1011},
  year={2022},
  publisher={IEEE}
}

@article{twinanda2017endonet,
  title={{EndoNet: a deep architecture for recognition tasks on laparoscopic videos}},
  author={Twinanda, Andru P and Shehata, Sherif and Mutter, Didier and Marescaux, Jacques and De Momi, Elena and Padoy, Nicolas},
  journal={IEEE Transactions on Medical Imaging},
  volume={36},
  number={1},
  pages={86--97},
  year={2017},
  publisher={IEEE}
}

@article{chen2024robotic,
  title={Robotic grasp detection using structure prior attention and multiscale features},
  author={Chen, Lu and Niu, Mingdi and Yang, Jing and Qian, Yuhua and Li, Zhuomao and Wang, Keqi and Yan, Tao and Huang, Panfeng},
  journal={IEEE Transactions on Systems, Man, and Cybernetics: Systems},
  year={2024},
  publisher={IEEE}
}

@article{zhao2024position,
  title={Position-aware pushing and grasping synergy with deep reinforcement learning in clutter},
  author={Zhao, Min and Zuo, Guoyu and Yu, Shuangyue and Gong, Daoxiong and Wang, Zihao and Sie, Ouattara},
  journal={CAAI Transactions on Intelligence Technology},
  volume={9},
  number={3},
  pages={738--755},
  year={2024},
  publisher={Wiley Online Library}
}

@article{kim2025srt,
  title={SRT-H: A hierarchical framework for autonomous surgery via language-conditioned imitation learning},
  author={Kim, Ji Woong and Chen, Juo-Tung and Hansen, Pascal and Shi, Lucy Xiaoyang and Goldenberg, Antony and Schmidgall, Samuel and Scheikl, Paul Maria and Deguet, Anton and White, Brandon M and Tsai, De Ru and others},
  journal={Science robotics},
  volume={10},
  number={104},
  pages={eadt5254},
  year={2025},
  publisher={American Association for the Advancement of Science}
}

@article{long2025surgical,
  title={Surgical embodied intelligence for generalized task autonomy in laparoscopic robot-assisted surgery},
  author={Long, Yonghao and Lin, Anran and Kwok, Derek Hang Chun and Zhang, Lin and Yang, Zhenya and Shi, Kejian and Song, Lei and Fu, Jiawei and Lin, Hongbin and Wei, Wang and others},
  journal={Science Robotics},
  volume={10},
  number={104},
  pages={eadt3093},
  year={2025},
  publisher={American Association for the Advancement of Science}
}

@inproceedings{chen2024worldafford,
  title={Worldafford: Affordance grounding based on natural language instructions},
  author={Chen, Changmao and Cong, Yuren and Kan, Zhen},
  booktitle={2024 IEEE 36th International Conference on Tools with Artificial Intelligence (ICTAI)},
  pages={822--828},
  year={2024},
  organization={IEEE}
}

@article{schmidgall2025will,
  title={Will your next surgeon be a robot? Autonomy and AI in robotic surgery},
  author={Schmidgall, Samuel and Opfermann, Justin D and Kim, Ji Woong and Krieger, Axel},
  journal={Science Robotics},
  volume={10},
  number={104},
  pages={eadt0187},
  year={2025},
  publisher={American Association for the Advancement of Science}
}

@article{knudsen2024clinical,
  title={Clinical applications of artificial intelligence in robotic surgery},
  author={Knudsen, J Everett and Ghaffar, Umar and Ma, Runzhuo and Hung, Andrew J},
  journal={Journal of robotic surgery},
  volume={18},
  number={1},
  pages={102},
  year={2024},
  publisher={Springer}
}

@article{kim2024surgical,
  title={Surgical robot transformer (srt): Imitation learning for surgical tasks},
  author={Kim, Ji Woong and Zhao, Tony Z and Schmidgall, Samuel and Deguet, Anton and Kobilarov, Marin and Finn, Chelsea and Krieger, Axel},
  journal={arXiv preprint arXiv:2407.12998},
  year={2024}
}

@inproceedings{moghani2025sufia,
  title={Sufia-bc: Generating high quality demonstration data for visuomotor policy learning in surgical subtasks},
  author={Moghani, Masoud and Nelson, Nigel and Ghanem, Mohamed and Diaz-Pinto, Andres and Hari, Kush and Azizian, Mahdi and Goldberg, Ken and Huver, Sean and Garg, Animesh},
  booktitle={2025 IEEE International Conference on Robotics and Automation (ICRA)},
  pages={4534--4541},
  year={2025},
  organization={IEEE}
}

@article{li2024review,
  title={A review of deep learning-based information fusion techniques for multimodal medical image classification},
  author={Li, Yihao and Daho, Mostafa El Habib and Conze, Pierre-Henri and Zeghlache, Rachid and Le Boit{\'e}, Hugo and Tadayoni, Ramin and Cochener, B{\'e}atrice and Lamard, Mathieu and Quellec, Gwenol{\'e}},
  journal={Computers in Biology and Medicine},
  volume={177},
  pages={108635},
  year={2024},
  publisher={Elsevier}
}

@article{ayobi2025pixel,
  title={Pixel-wise recognition for holistic surgical scene understanding},
  author={Ayobi, Nicol{\'a}s and Rodr{\'\i}guez, Santiago and P{\'e}rez, Alejandra and Hern{\'a}ndez, Isabela and Aparicio, Nicol{\'a}s and Dessevres, Eug{\'e}nie and Pe{\~n}a, Sebasti{\'a}n and Santander, Jessica and Caicedo, Juan Ignacio and Fern{\'a}ndez, Nicol{\'a}s and others},
  journal={Medical Image Analysis},
  pages={103726},
  year={2025},
  publisher={Elsevier}
}

@inproceedings{wang2022autolaparo,
  title={Autolaparo: A new dataset of integrated multi-tasks for image-guided surgical automation in laparoscopic hysterectomy},
  author={Wang, Ziyi and Lu, Bo and Long, Yonghao and Zhong, Fangxun and Cheung, Tak-Hong and Dou, Qi and Liu, Yunhui},
  booktitle={International Conference on Medical Image Computing and Computer-Assisted Intervention},
  pages={486--496},
  year={2022},
  organization={Springer}
}

@article{nwoye2022rendezvous,
  title={Rendezvous: Attention mechanisms for the recognition of surgical action triplets in endoscopic videos},
  author={Nwoye, Chinedu Innocent and Yu, Tong and Gonzalez, Cristians and Seeliger, Barbara and Mascagni, Pietro and Mutter, Didier and Marescaux, Jacques and Padoy, Nicolas},
  journal={Medical Image Analysis},
  volume={78},
  pages={102433},
  year={2022},
  publisher={Elsevier}
}

@article{allan20202018,
  title={2018 robotic scene segmentation challenge},
  author={Allan, Max and Kondo, Satoshi and Bodenstedt, Sebastian and Leger, Stefan and Kadkhodamohammadi, Rahim and Luengo, Imanol and Fuentes, Felix and Flouty, Evangello and Mohammed, Ahmed and Pedersen, Marius and others},
  journal={arXiv preprint arXiv:2001.11190},
  year={2020}
}

@article{maier2021heidelberg,
  title={Heidelberg colorectal data set for surgical data science in the sensor operating room},
  author={Maier-Hein, Lena and Wagner, Martin and Ross, Tobias and Reinke, Annika and Bodenstedt, Sebastian and Full, Peter M and Hempe, Hellena and Mindroc-Filimon, Diana and Scholz, Patrick and Tran, Thuy Nuong and others},
  journal={Scientific data},
  volume={8},
  number={1},
  pages={101},
  year={2021},
  publisher={Nature Publishing Group UK London}
}

@article{bawa2021saras,
  title={The saras endoscopic surgeon action detection (esad) dataset: Challenges and methods},
  author={Bawa, Vivek Singh and Singh, Gurkirt and KapingA, Francis and Skarga-Bandurova, Inna and Oleari, Elettra and Leporini, Alice and Landolfo, Carmela and Zhao, Pengfei and Xiang, Xi and Luo, Gongning and others},
  journal={arXiv preprint arXiv:2104.03178},
  year={2021}
}

@article{lavanchy2024challenges,
  title={Challenges in multi-centric generalization: phase and step recognition in Roux-en-Y gastric bypass surgery},
  author={Lavanchy, Jo{\"e}l L and Ramesh, Sanat and Dall’Alba, Diego and Gonzalez, Cristians and Fiorini, Paolo and M{\"u}ller-Stich, Beat P and Nett, Philipp C and Marescaux, Jacques and Mutter, Didier and Padoy, Nicolas},
  journal={International journal of computer assisted radiology and surgery},
  volume={19},
  number={11},
  pages={2249--2257},
  year={2024},
  publisher={Springer}
}

@article{schoeffmann2018video,
  title={Video retrieval in laparoscopic video recordings with dynamic content descriptors},
  author={Schoeffmann, Klaus and Husslein, Heinrich and Kletz, Sabrina and Petscharnig, Stefan and Muenzer, Bernd and Beecks, Christian},
  journal={Multimedia Tools and Applications},
  volume={77},
  number={13},
  pages={16813--16832},
  year={2018},
  publisher={Springer}
}

@inproceedings{zhan2025tracking,
  title={Tracking everything in robotic-assisted surgery},
  author={Zhan, Bohan and Zhao, Wang and Fang, Yi and Du, Bo and Vasconcelos, Francisco and Stoyanov, Danail and Elson, Daniel S and Huang, Baoru},
  booktitle={2025 IEEE International Conference on Robotics and Automation (ICRA)},
  pages={1--7},
  year={2025},
  organization={IEEE}
}

\end{document}